\documentclass[journal]{IEEEtran}
\usepackage{graphicx}
\usepackage{amsmath}
\usepackage{amssymb}
\usepackage{booktabs}
\usepackage{multirow}
\usepackage{xcolor}
\usepackage{hyperref}
\usepackage{tabularx}

\title{Cross-Architecture Knowledge Distillation from a Vision Foundation Model to a Lightweight Visual State Space Model for Tea Leaf Disease Classification}

\author{Zibo Zhou, Zongsen Qiu, Rui Chen, Yujie Yao, Yue Zhou, Jianjun Wang}

\begin{document}
\maketitle

\begin{abstract}
Automated tea leaf disease classification supports precision agriculture, yet deploying accurate models on edge devices remains challenging under tight compute budgets. Self-supervised vision foundation models such as DINOv2 provide strong features but are too large for field deployment, while lightweight models trained from scratch on small agricultural datasets often underfit. We study cross-architecture knowledge distillation (KD) from a fine-tuned DINOv2 teacher (a Vision Transformer) to a compact bidirectional Visual State Space Model (LVSSM) student, a direction that is comparatively underexplored because the two architectures use fundamentally different token-mixing mechanisms. We first identify and fix two training-stability problems that otherwise prevent our from-scratch SSM student from learning on limited data: a single large patch-embedding convolution and a fusion layer that severs the residual path. With a progressive convolutional stem and a gated bidirectional selective-scan block, the 4.45M-parameter student trains stably. Averaged over three seeds, temperature-scaled logit distillation raises test accuracy from 92.32$\pm$2.14\% (standalone) to 95.41$\pm$1.17\% (best single run 96.20\%, macro-F1 94.45\%), a +3.09 percentage-point mean gain that also reduces run-to-run variance, while the model uses 5.0$\times$ fewer parameters than the 22M-parameter teacher and retains 98.3\% of its accuracy. Through controlled ablations that hold the distillation weight fixed, we find that adding intermediate feature-alignment losses \emph{reduces} accuracy, so simple logit-level KD is the strongest configuration for this ViT$\rightarrow$SSM transfer. A fair from-scratch comparison shows the gain is specific to students that start below the teacher: the same recipe helps the SSM but not from-scratch CNNs that already match the teacher. We report per-class metrics, confusion matrices, bootstrap confidence intervals, and FLOPs/latency measurements, and we discuss the limitations of the study, including its single-dataset scope and the use of a simplified (non-official) SSM implementation.
\end{abstract}


\section{Introduction}

Tea is among the most widely consumed beverages worldwide, and its yield and quality depend on timely detection of foliar diseases. Manual inspection is labor-intensive and subjective, which motivates automated image-based classification. While deep learning has advanced plant disease recognition substantially, deploying high-accuracy models on the resource-constrained edge devices typical of agricultural settings remains an open problem.

Self-supervised vision foundation models, particularly DINOv2~\cite{oquab2023dinov2}, achieve strong transfer accuracy across visual tasks. However, even their smaller variants ($\sim$22M parameters) are expensive for embedded inference. Conversely, compact architectures trained from scratch on small agricultural datasets frequently fail to learn discriminative features and generalize poorly.

State Space Models (SSMs), and the selective-scan Mamba formulation~\cite{gu2023mamba} in particular, offer linear-time sequence modeling and have been adapted to vision by Vision Mamba (Vim)~\cite{zhu2024vision} and VMamba~\cite{liu2024vmamba}. Their favorable complexity makes SSMs attractive for edge deployment. However, training vision SSMs from scratch on small datasets is difficult: the bidirectional scan is prone to unstable gradient flow, and naive patch embeddings provide little local inductive bias.

Knowledge distillation (KD)~\cite{hinton2015distilling} is a natural way to transfer the knowledge of a large teacher into a compact student. Most KD literature targets same-family transfer (CNN$\rightarrow$CNN, ViT$\rightarrow$ViT), where teacher and student share token-mixing mechanisms and intermediate representations can be aligned directly. \emph{Cross-architecture} distillation between families with different inductive biases --- here, a ViT teacher (global self-attention over patches) and an SSM student (sequential state propagation over a scanned patch order) --- is less understood, and it is unclear whether feature-level alignment helps or hurts.

In this work we make the following contributions:
\begin{enumerate}
    \item We diagnose two concrete failure modes that prevent a from-scratch bidirectional visual SSM from learning on a small dataset, and we fix them with a progressive convolutional stem and a gated-residual bidirectional selective-scan block. The resulting 4.45M-parameter student (which we call LVSSM, for Lightweight Visual State Space Model) trains stably to 92.16\% test accuracy without any pretraining.
    \item We study cross-architecture KD from a fine-tuned DINOv2 teacher to this SSM student. Averaged over three seeds, logit-level distillation improves student accuracy by +3.09 points (92.32\%$\rightarrow$95.41\%; best single run 96.20\%) and reduces variance, demonstrating that dark knowledge transfers effectively even across strongly dissimilar architectures.
    \item Through controlled ablations with a fixed CE/KD weighting, we find that adding intermediate feature-alignment losses does not improve over logit-only KD at this scale, and we analyze why. We report a complete evaluation (per-class metrics, confusion matrices, bootstrap confidence intervals, FLOPs and GPU/CPU latency) and discuss the study's limitations honestly.
\end{enumerate}

\section{Related Work}

\subsection{Deep Learning for Plant Disease Classification}
CNN-based transfer learning from ImageNet-pretrained backbones (ResNet~\cite{he2016deep}, EfficientNet~\cite{tan2019efficientnet}, MobileNet~\cite{howard2019searching}) is the dominant paradigm for plant disease recognition, and large curated datasets such as PlantVillage~\cite{mohanty2016using} have driven progress~\cite{ferentinos2018deep}. These approaches achieve high accuracy but depend on large-scale external pretraining and are typically benchmarked without explicit attention to deployment cost. Tea leaf disease datasets are comparatively small and class-imbalanced, which makes both from-scratch training and fair efficiency comparison more delicate.

\subsection{Vision State Space Models}
Mamba~\cite{gu2023mamba} introduced input-dependent (selective) state space parameters, giving linear-time sequence modeling with a hardware-aware scan. Vision Mamba~\cite{zhu2024vision} and VMamba~\cite{liu2024vmamba} adapt selective scanning to images by serializing patches and scanning them bidirectionally or along multiple spatial orders. These models are efficient but, in our experience, require careful architectural choices (stem design, residual structure) to train stably on small datasets. Our student is a deliberately simplified, self-contained SSM block inspired by this line of work rather than the official \texttt{mamba-ssm} kernel; we therefore name it LVSSM to avoid overstating equivalence with Vim/VMamba.

\subsection{Knowledge Distillation}
KD~\cite{hinton2015distilling} transfers a teacher's softened output distribution to a student. Feature-based extensions align intermediate representations, and review-based methods~\cite{chen2022distilling} match features across multiple stages. Most such methods assume architecturally compatible teacher--student pairs. Cross-architecture transfer --- e.g., ViT$\rightarrow$CNN or, as here, ViT$\rightarrow$SSM --- is less studied; the central difficulty is that intermediate features encode information in structurally different coordinate systems (spatial attention maps vs. scan-order state trajectories), so direct feature alignment may inject noise rather than signal. Our ablations speak directly to this question.

\section{Methodology}

\subsection{Problem Formulation}
Given a teacher $T$ (fine-tuned DINOv2) and a student $S$ (LVSSM), we train $S$ to approximate $T$'s predictive distribution while using far fewer parameters. Let $\mathcal{D}=\{(x_i,y_i)\}_{i=1}^N$ be the labeled training set of tea leaf images.

\subsection{Teacher Model: Fine-tuned DINOv2}
We use DINOv2-Small (22.06M parameters) as the teacher, fine-tuned end-to-end on the tea disease dataset with a linear classification head over the class token. Under our unified evaluation pipeline the teacher attains 97.86\% test accuracy and 96.76\% macro-F1, serving as a strong reference.

\subsection{Student Model: Lightweight Visual SSM (LVSSM)}
Our student (4,454,982 parameters) addresses two training-stability failure modes we observed when training a bidirectional visual SSM from scratch.

\subsubsection{Diagnosed failure modes}
A baseline design using (i) a single $16\times16$ patch-embedding convolution and (ii) a bidirectional block whose forward/backward outputs are merged by a linear ``fusion'' layer followed by LayerNorm exhibited severe vanishing gradients: measured gradient norms decayed from $\sim$0.85 at the first block to $\sim$0.03 at the second and to near zero in deeper blocks, and the model stayed at chance accuracy ($\sim$1/6). We attribute this to (i) the single large convolution providing no local inductive bias before sequential scanning, and (ii) the fusion layer routing \emph{all} information through a non-identity transform, which destroys the residual pathway.

\subsubsection{Progressive Convolutional Stem}
We replace the single convolution with a 4-layer progressive stem of stride-2 convolutions,
\begin{equation}
    \text{ConvStem}: \mathbb{R}^{3\times H\times W}\rightarrow\mathbb{R}^{d\times\frac{H}{16}\times\frac{W}{16}},
\end{equation}
with channel widths $d/4\rightarrow d/2\rightarrow d\rightarrow d$, each followed by BatchNorm and GELU. This provides local feature extraction before sequential processing, mirroring the stems used in successful vision SSMs.

\subsubsection{Gated-Residual Bidirectional Selective Scan}
Each block scans the sequence forward and backward and combines the two directions through a learned gate while preserving an explicit identity path:
\begin{equation}
    \mathbf{c}=[\text{SSM}_{\text{fwd}}(\mathbf{x});\ \text{SSM}_{\text{bwd}}(\text{flip}(\mathbf{x}))],
\end{equation}
\begin{equation}
    \mathbf{g}=\sigma(W_g\,\mathbf{c}),\qquad
    \mathbf{o}=\mathbf{x}+\mathbf{g}\odot W_p\,\mathbf{c}.
\end{equation}
The gate $\mathbf{g}$ learns how much bidirectional context to inject, while the additive residual guarantees gradient flow through the identity path. This single change restored stable training in all our runs.

\subsubsection{Configuration}
The student stacks 6 gated bidirectional SSM blocks with embedding dimension $d=192$, state dimension 16, expansion factor 2, and convolution kernel size 4, producing 196 tokens ($14\times14$) at $224\times224$ input and 4,454,982 parameters in total.

\subsection{Distillation Framework}

\subsubsection{Logit-level Distillation}
The primary objective is temperature-scaled soft-target matching:
\begin{equation}
    \mathcal{L}_\text{KD}=\tau^2\cdot\text{KL}\!\left(\sigma(z_s/\tau)\ \|\ \sigma(z_t/\tau)\right),
\end{equation}
where $z_s,z_t$ are student/teacher logits, $\tau$ is the temperature, and $\sigma$ is the softmax. We use $\tau=2.0$; the $\tau^2$ factor keeps the KD gradient magnitude comparable to the cross-entropy term.

\subsubsection{Combined Objective}
The total loss combines ground-truth cross-entropy and KD:
\begin{equation}
    \mathcal{L}=\alpha_\text{CE}\,\mathcal{L}_\text{CE}+\alpha_\text{KD}\,\mathcal{L}_\text{KD},
\end{equation}
with $\alpha_\text{CE}=\alpha_\text{KD}=0.5$ in the main configuration. The ablation study (Section~\ref{sec:ablation}) additionally considers two optional feature-alignment terms: a scan-attention alignment loss $\mathcal{L}_\text{SAA}$ that matches a pooled teacher attention signal to a student scan-state summary, and a progressive feature-distillation loss $\mathcal{L}_\text{Prog}$ that aligns projected intermediate features at several depths.

\subsection{Training Strategy}
We use AdamW (weight decay 0.05), a peak learning rate of $5\times10^{-4}$, linear warmup (start factor 0.01) followed by cosine annealing to $10^{-6}$, and gradient clipping at norm 1.0. Augmentation comprises random resized crop, horizontal/vertical flips, color jitter, and mild rotation. The distilled students use batch size 12 (bounded by the bidirectional-scan memory footprint); the standalone student uses batch size 24. All models train for 30 epochs.

\section{Experiments}

\subsection{Dataset and Setup}
We evaluate on a real-world tea leaf disease dataset (Roboflow, CC~BY~4.0) with six classes: Algal Leaf Spot, Brown Blight, Gray Blight, Healthy, Helopeltis, and Red Leaf Spot. The dataset provides a group-aware split by source photograph, which we verified: cross-split overlap of source-photo identifiers is exactly zero between train, validation, and test, so there is no data leakage. After converting the original detection-format annotations to classification labels (discarding 79 images with empty annotation files, i.e.\ no bounding box), the splits contain 4{,}259 training, 421 validation, and 421 test images at $224\times224$. The test set is class-imbalanced (Healthy: 192; the other five classes: 41--50), which is why we emphasize macro-F1 and per-class metrics rather than accuracy alone. All experiments run on a single NVIDIA RTX 4090.

\subsection{Implementation Details}
The teacher (DINOv2-Small) is fine-tuned for 30 epochs at batch size 32. CNN baselines (ResNet18, EfficientNet-B0, MobileNetV3-Small) are ImageNet-pretrained and fine-tuned for 30 epochs. The LVSSM student is trained for 30 epochs as described above. To ensure comparability, every model is evaluated with a single shared pipeline that computes accuracy, macro/weighted-F1, per-class precision/recall/F1, confusion matrices, bootstrap 95\% confidence intervals (2{,}000 resamples), FLOPs (via \texttt{ptflops}), and GPU/CPU single-image latency.

\subsection{Main Results}

\begin{table}[t]
\centering
\caption{Main comparison on the tea leaf disease test set (421 images), all under one evaluation pipeline. Values are best single-run (seed 42); student multi-seed means are in Table~\ref{tab:multiseed}. Best student in bold.}
\label{tab:main_results}
\begin{tabular}{lcccc}
\toprule
\textbf{Model} & \textbf{Params} & \textbf{Acc (\%)} & \textbf{Macro-F1 (\%)} & \textbf{Pretrain} \\
 & \textbf{(M)} & & & \\
\midrule
DINOv2-S (Teacher) & 22.06 & 97.86 & 96.76 & SSL-142M \\
\midrule
ResNet18 & 11.18 & 98.34 & 97.52 & ImageNet \\
EfficientNet-B0 & 4.02 & 98.57 & 97.89 & ImageNet \\
MobileNetV3-S & 1.52 & 98.10 & 97.12 & ImageNet \\
\midrule
LVSSM (standalone) & 4.45 & 92.16 & 89.82 & None \\
LVSSM + Full distill & 4.45 & 95.49 & 93.79 & None \\
LVSSM + Logit KD & 4.45 & \textbf{96.20} & \textbf{94.45} & None \\
\bottomrule
\end{tabular}
\end{table}

Table~\ref{tab:main_results} summarizes the main results. Three observations stand out.

\textbf{Cross-architecture KD is effective.} For the best single run (seed 42), logit distillation improves the SSM student from 92.16\% to 96.20\% accuracy and from 89.82\% to 94.45\% macro-F1; averaged over three seeds the gain is +3.09 accuracy / +3.44 macro-F1 points (Table~\ref{tab:multiseed}). Either way, the teacher's softened outputs transfer useful ``dark knowledge'' despite the ViT$\rightarrow$SSM architecture gap. Table~\ref{tab:main_results} reports best-run values for each model; multi-seed means for the student are in Table~\ref{tab:multiseed}.

\textbf{The efficiency tradeoff is favorable.} The distilled student uses 5.0$\times$ fewer parameters than the teacher and retains 98.3\% of its accuracy (96.20 vs.\ 97.86).

\textbf{Pretraining still matters.} ImageNet-pretrained CNNs (98.1--98.6\%) lead the from-scratch SSM student. This gap reflects large-scale external pretraining data unavailable to the student rather than architectural inferiority; the distilled SSM, trained only on 4{,}259 tea images, comes within $\sim$2.4 points of EfficientNet-B0.

\begin{figure}[t]
\centering
\includegraphics[width=\columnwidth]{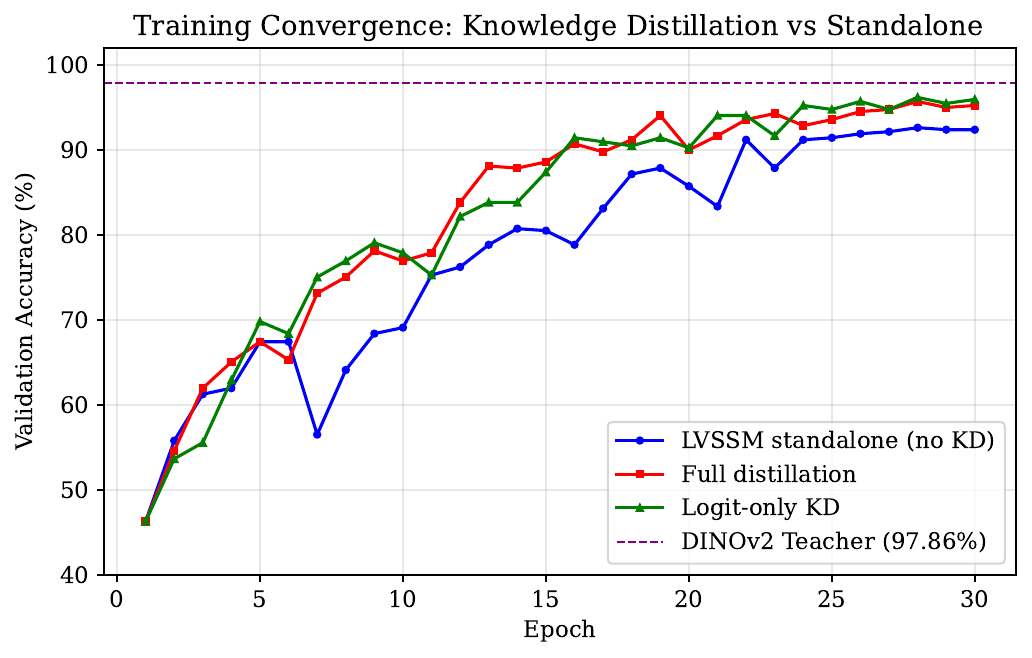}
\caption{Validation accuracy over training. Distillation converges faster and to a higher plateau than standalone training.}
\label{fig:training_curves}
\end{figure}

\begin{figure}[t]
\centering
\includegraphics[width=\columnwidth]{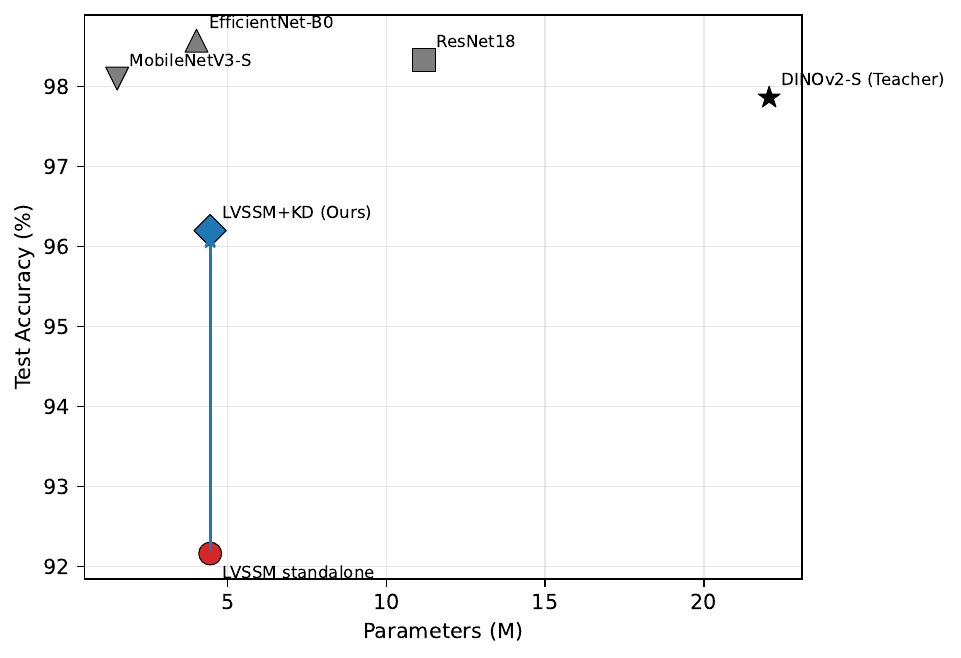}
\caption{Accuracy vs.\ parameters. Distillation lifts the SSM student (arrow) without any external pretraining data.}
\label{fig:params_accuracy}
\end{figure}

\subsection{Per-Class Analysis and Error Structure}

\begin{table}[t]
\centering
\caption{Per-class F1 (\%) on the test set. Support in parentheses.}
\label{tab:per_class}
\begin{tabular}{lccc}
\toprule
\textbf{Class (support)} & \textbf{Standalone} & \textbf{Logit KD} & \textbf{Teacher} \\
\midrule
Algal Leaf Spot (42) & 96.39 & 95.35 & 98.82 \\
Brown Blight (41)    & 85.00 & 87.80 & 91.57 \\
Gray Blight (47)     & 86.32 & 88.42 & 92.47 \\
Healthy (192)        & 96.62 & 99.22 & 99.74 \\
Helopeltis (49)      & 84.21 & 96.91 & 98.97 \\
Red Leaf Spot (50)   & 90.38 & 98.99 & 98.99 \\
\midrule
\textbf{Macro-F1}    & 89.82 & 94.45 & 96.76 \\
\bottomrule
\end{tabular}
\end{table}

\begin{figure}[t]
\centering
\includegraphics[width=0.85\columnwidth]{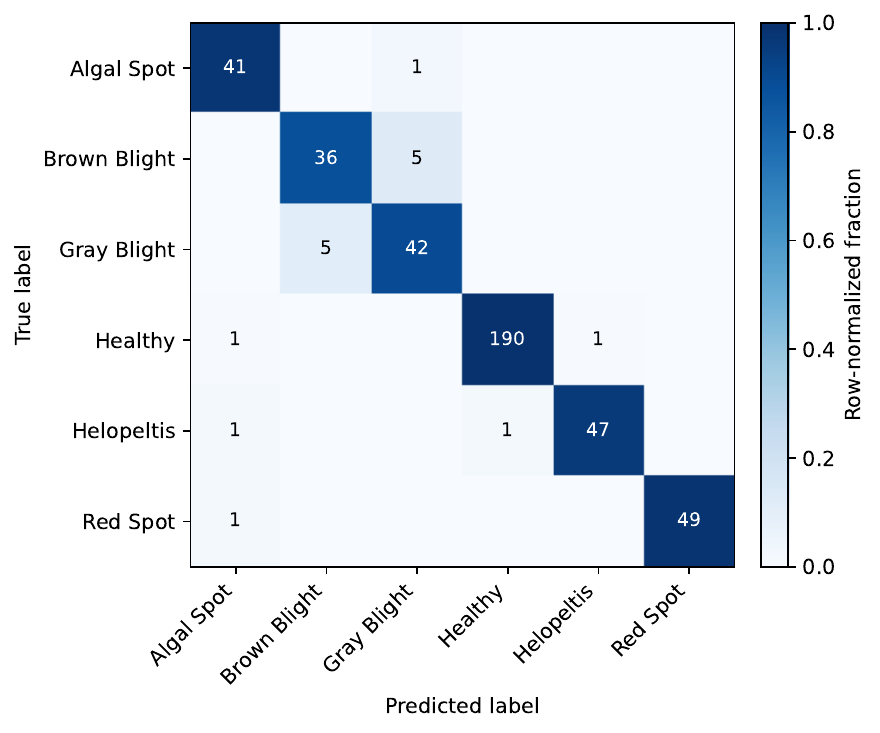}
\caption{Confusion matrix of the distilled LVSSM (logit KD). The main residual error is Brown Blight $\leftrightarrow$ Gray Blight.}
\label{fig:confusion}
\end{figure}

Table~\ref{tab:per_class} shows that distillation helps most on the classes the standalone student handled worst: Helopeltis (+12.7 F1) and Red Leaf Spot (+8.6 F1). The dominant residual error, visible in the confusion matrix (Fig.~\ref{fig:confusion}), is mutual confusion between Brown Blight and Gray Blight (5 misclassifications in each direction) --- two visually similar necrotic-lesion diseases that the teacher itself also confuses (Table~\ref{tab:per_class}). This suggests the error is intrinsic to the visual similarity of the two diseases rather than a distillation artifact.

\begin{figure}[t]
\centering
\includegraphics[width=\columnwidth]{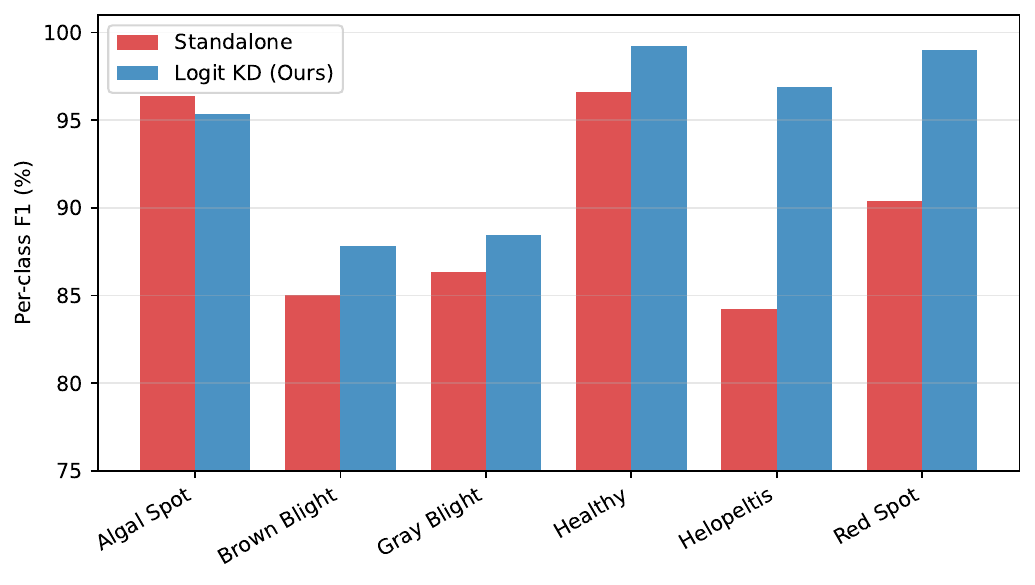}
\caption{Per-class F1 before and after distillation. Gains concentrate on the hardest standalone classes.}
\label{fig:per_class_f1}
\end{figure}

\subsection{Ablation: Does Feature Alignment Help?}
\label{sec:ablation}

We consider two optional feature-alignment terms on top of logit KD: a scan-attention alignment loss ($\mathcal{L}_\text{SAA}$) and a progressive feature-distillation loss ($\mathcal{L}_\text{Prog}$). A naive ablation that redistributes a fixed loss budget (lowering $\alpha_\text{KD}$ to make room for the new terms) confounds two effects --- the feature term's contribution and the weaker KD signal. We therefore run a \emph{controlled} ablation that holds $\alpha_\text{CE}=\alpha_\text{KD}=0.5$ fixed and \emph{adds} each feature term at weight 0.5, changing nothing about the primary objective.

\begin{table}[t]
\centering
\caption{Controlled ablation: CE and KD weights held fixed at 0.5, feature-alignment terms \emph{added} on top (not traded against KD). Identical schedule, $\tau=2.0$, 30 epochs.}
\label{tab:ablation}
\begin{tabular}{lcccc|c}
\toprule
\textbf{Config} & $\alpha_\text{CE}$ & $\alpha_\text{KD}$ & $\alpha_\text{SAA}$ & $\alpha_\text{Prog}$ & \textbf{Acc (\%)} \\
\midrule
No distillation & 1.0 & 0.0 & 0.0 & 0.0 & 92.16 \\
\midrule
Logit-only & 0.5 & 0.5 & 0.0 & 0.0 & \textbf{96.20} \\
+ SAA & 0.5 & 0.5 & 0.5 & 0.0 & 94.77 \\
+ Progressive & 0.5 & 0.5 & 0.0 & 0.5 & 95.49 \\
+ SAA + Prog & 0.5 & 0.5 & 0.5 & 0.5 & 95.01 \\
\bottomrule
\end{tabular}
\end{table}

Table~\ref{tab:ablation} gives an unambiguous answer: even with the KD signal held at full strength, adding feature-alignment losses \emph{reduces} accuracy (96.20\%$\rightarrow$94.77--95.49\%). This rules out the ``diluted KD weight'' explanation --- the feature terms are not merely neutral, they actively interfere. We attribute this to a representational mismatch: the SAA loss forces the student's scan-state summary toward the teacher's patch-attention geometry, and the progressive loss aligns intermediate features across a ViT and an SSM whose depth-wise representations are structurally different. Constraining the SSM to imitate ViT-shaped intermediate features conflicts with the representations the scan mechanism would otherwise learn. Logit-only KD, which constrains only the output distribution, avoids this conflict and is the strongest configuration.

\begin{figure}[t]
\centering
\includegraphics[width=\columnwidth]{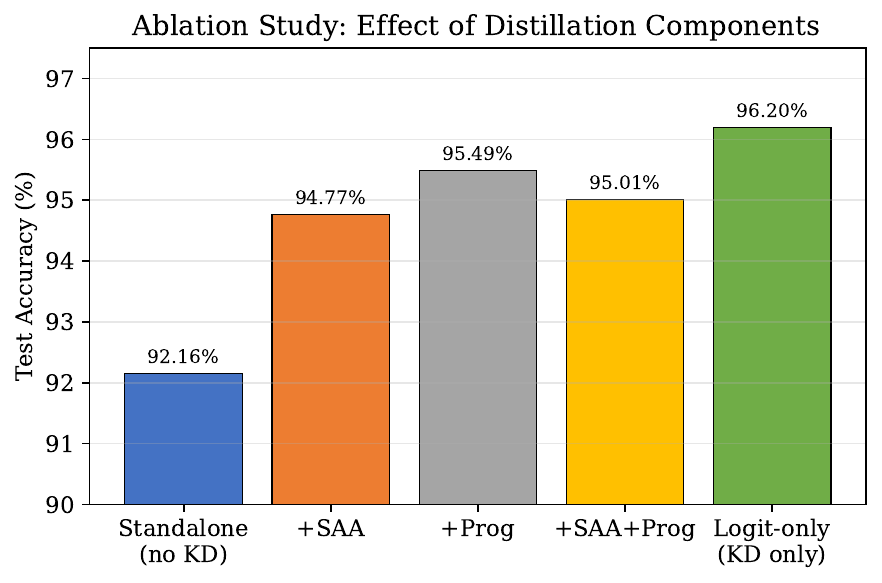}
\caption{Controlled ablation. With KD held at full weight, adding feature alignment reduces accuracy; logit-only KD is strongest.}
\label{fig:ablation}
\end{figure}

\subsection{Is Cross-Architecture KD Special to the SSM Student?}
\label{sec:fair}
A natural question is whether the distillation gain reflects something specific about the SSM student or whether the same teacher would help any compact from-scratch model. To test this, we distill the same DINOv2 teacher into two CNN students (MobileNetV3-Small, EfficientNet-B0) trained \emph{from scratch} (no ImageNet), using the identical logit-KD recipe ($\tau=2.0$, $\alpha_\text{CE}=\alpha_\text{KD}=0.5$). This is the apples-to-apples comparison: all three students start from random initialization.

\begin{table}[t]
\centering
\caption{Fair cross-architecture distillation: same teacher, all students trained from scratch (no pretraining), identical logit-KD recipe.}
\label{tab:fair}
\begin{tabular}{lccc}
\toprule
\textbf{Student (from scratch)} & \textbf{Standalone} & \textbf{+ Logit KD} & $\Delta$ \\
\midrule
MobileNetV3-Small & 96.44 & 95.25 & $-$1.19 \\
EfficientNet-B0 & 97.15 & 97.39 & +0.24 \\
LVSSM (ours) & 92.32 & 95.41 & \textbf{+3.09} \\
\bottomrule
\end{tabular}
\end{table}

Table~\ref{tab:fair} shows a clear pattern: distillation from this teacher yields a large gain for the SSM student (+3.09 points, mean over seeds) but is neutral-to-negative for the CNN students ($-$1.19 and +0.24). The explanation is that the from-scratch CNNs already reach 96.4--97.2\% standalone --- at or above the teacher's own 97.86\% --- so there is little useful ``dark knowledge'' left to transfer, and a not-strictly-stronger teacher can even mislead them. The SSM student, by contrast, starts well below the teacher and has substantial headroom. This clarifies the scope of our claim: cross-architecture KD is most valuable precisely when the compact student would otherwise underperform the teacher, as is the case for from-scratch vision SSMs on small data.

\subsection{Efficiency and Complexity}

\begin{table}[t]
\centering
\caption{Complexity and latency (single $224^2$ image; RTX 4090 for GPU, server CPU for CPU). FLOPs via \texttt{ptflops}.}
\label{tab:efficiency}
\begin{tabular}{lcccc}
\toprule
\textbf{Model} & \textbf{Params} & \textbf{GFLOPs} & \textbf{GPU} & \textbf{CPU} \\
 & \textbf{(M)} & & \textbf{(ms)} & \textbf{(ms)} \\
\midrule
DINOv2-S (Teacher) & 22.06 & 11.06 & 5.07 & 102.4 \\
ResNet18 & 11.18 & 3.65 & 1.42 & 53.6 \\
EfficientNet-B0 & 4.02 & 0.78 & 4.55 & 43.3 \\
MobileNetV3-S & 1.52 & 0.11 & 1.96 & 5.2 \\
LVSSM (Ours) & 4.45 & 2.19 & 10.05 & 1410 \\
\bottomrule
\end{tabular}
\end{table}

Table~\ref{tab:efficiency} reports the efficiency picture honestly. On parameters the distilled LVSSM (4.45M) is 5.0$\times$ smaller than the teacher. However, its measured latency is \emph{not} competitive: the pure-PyTorch bidirectional selective scan runs at 10.05\,ms on GPU and 1410\,ms on CPU, far slower than the CNN baselines, because our simplified SSM does not use the hardware-aware fused scan kernel of the official Mamba implementation. Parameter count and FLOPs (2.19 GFLOPs) therefore understate real inference cost. Closing this gap requires a fused scan kernel and is important future work before any edge-deployment claim.

\subsection{Multi-Seed Stability}
Because a single-seed gain could partly reflect initialization noise, we repeat the standalone and logit-KD configurations across three random seeds (42, 123, 456) and report mean\,$\pm$\,std (Table~\ref{tab:multiseed}). The distillation gain is robust: averaged over seeds, logit KD improves accuracy by +3.09 points (92.32$\rightarrow$95.41) and macro-F1 by +3.44 points (90.19$\rightarrow$93.63). Distillation also \emph{reduces} run-to-run variance (accuracy std 2.14$\rightarrow$1.17), consistent with the teacher's soft targets providing a more stable optimization signal than one-hot labels alone.

\begin{table}[t]
\centering
\caption{Multi-seed results (seeds 42, 123, 456). Mean\,$\pm$\,std over three runs.}
\label{tab:multiseed}
\begin{tabular}{lcc}
\toprule
\textbf{Config} & \textbf{Accuracy (\%)} & \textbf{Macro-F1 (\%)} \\
\midrule
Standalone & 92.32\,$\pm$\,2.14 & 90.19\,$\pm$\,2.50 \\
Logit KD   & \textbf{95.41\,$\pm$\,1.17} & \textbf{93.63\,$\pm$\,1.45} \\
\midrule
$\Delta$ (KD $-$ standalone) & +3.09 & +3.44 \\
\bottomrule
\end{tabular}
\end{table}

Note that the best single logit-KD run (seed 42, 96.20\%) and the mean (95.41\%) differ by less than one standard deviation; we report both, and use the mean for all cross-model comparisons where variance matters.

\section{Discussion}

\textbf{Why does logit KD transfer across such different architectures?} Logit-level KD only constrains the student's output distribution, leaving it free to develop internal representations suited to sequential scanning. Feature-level alignment, by contrast, would ask the SSM to reproduce the teacher's patch-attention geometry, which has no natural analogue in scan-order state trajectories. This is consistent with our ablation and with the broader intuition that output-space transfer is more architecture-agnostic than feature-space transfer.

\textbf{Practical guidance.} For distillation across dissimilar families (ViT$\rightarrow$SSM, and likely CNN$\rightarrow$SSM), a well-tuned logit-only KD is a strong default; our controlled ablation shows that adding ViT-style feature alignment hurts even when the KD weight is held at full strength, so it should not be adopted without careful per-task validation against a logit-only baseline. Equally important is \emph{when} to distill at all: our fair-baseline experiment (Section~\ref{sec:fair}) shows the gain is large only when the from-scratch student would otherwise fall below the teacher (the SSM), and vanishes when it already matches the teacher (the CNNs). Cross-architecture KD is therefore best viewed as a tool for closing a student--teacher gap, not a universal accuracy boost.

\textbf{Limitations.}
(1) \emph{Single dataset.} We evaluate on one 6-class tea dataset ($\sim$4K training images); the balance between logit and feature KD may shift with scale or task complexity.
(2) \emph{Simplified SSM.} Our student is a self-contained SSM approximation, not the official \texttt{mamba-ssm} kernel; accordingly we name it LVSSM and do not claim parity with Vim/VMamba, and its unfused scan is slow at inference (Table~\ref{tab:efficiency}).
(3) \emph{Grouping granularity.} The split is group-aware by source photograph (verified leakage-free), but we cannot rule out near-duplicate specimens photographed in separate sessions.
(4) \emph{No edge hardware.} We report GPU/CPU latency on server hardware, not on embedded devices (e.g., Jetson); real deployment suitability requires on-device measurement and a fused kernel.
(5) \emph{Teacher ceiling.} Under our unified pipeline the fine-tuned teacher reaches 97.86\%, below the ImageNet-pretrained CNNs, so the student inherits a teacher that is not the strongest available model.

\section{Conclusion}
We presented a cross-architecture knowledge distillation study transferring knowledge from a DINOv2 teacher to a lightweight bidirectional Visual State Space Model for tea leaf disease classification. We identified and fixed two failure modes that block from-scratch SSM training on small data, and showed that simple logit-level distillation lifts the 4.45M-parameter student by +3.09 points on average over three seeds (to 95.41\%; best run 96.20\%, 94.45\% macro-F1) --- retaining 98.3\% of the teacher's accuracy with 5.0$\times$ fewer parameters.Controlled ablations, with the KD weight held at full strength, show that adding ViT-style feature-alignment losses \emph{reduces} accuracy, so logit KD alone is the strongest configuration for this ViT$\rightarrow$SSM transfer. A fair-baseline comparison further shows the distillation gain is specific to students that start below the teacher: the same recipe helps the from-scratch SSM (+3.09) but not from-scratch CNNs that already match the teacher. We release complete per-class metrics, confusion matrices, confidence intervals, and complexity measurements, and we identify a fused selective-scan kernel and multi-dataset validation as the key steps toward practical edge deployment.

\section*{Acknowledgment}
This work was supported by the Innovation Training Program for College Students in Sichuan Province (No.202510626013).
\bibliographystyle{IEEEtran}
\bibliography{references}

\end{document}